\documentclass{article}
\usepackage{authblk}
\usepackage{hyperref}
\usepackage{xurl}
\hypersetup{colorlinks,allcolors=black}
\newcommand{\aparef}{\noindent\hangindent=0.5in\hangafter=1}
\usepackage{wrapfig}
\usepackage{graphicx} 
\usepackage[margin=1in]{geometry}

\title{From FAIR to AI-FAIR: Metadata-First Infrastructure for AI-Enabled Science}

\begin{document}

\author[1]{Elizabeth G. Campolongo\footnote{Equal Contribution: campolongo.4@osu.edu, esokol@battelleecology.org}}
\author[2]{Eric R. Sokol*}
\author[3]{Marty Downs} 
\author[4]{Gina Glenne}
\author[5]{Gregory E. Maurer}
\author[6]{Sydne Record} 
\author[7]{Dustin R. Rubenstein}
\author[1]{Matthew J. Thompson}
\author[2]{Christine Laney}
\author[8]{Hilmar Lapp}
\author[9]{Max Tarjan}
\author[1]{Tanya Berger-Wolf}

\affil[1]{Department of Computer Science and Engineering, The
Ohio State University}
\affil[2]{National Ecological
Observatory Network (NEON), Battelle}
\affil[3]{Long Term Ecological Research Network, National Center for Ecological Analysis and Synthesis, UC Santa Barbara}
\affil[4]{US Fish and Wildlife Service}
\affil[5]{New Mexico State University}
\affil[6]{University of Maine \& Maine Agricultural and Forest Experiment Station}
\affil[7]{Ecology, Evolution and Environmental Biology, Columbia University, NSCORE}
\affil[8]{Neuromatch}
\affil[9]{NatureServe}

\date{November 2026}

\maketitle

\begin{abstract}
Artificial intelligence (AI) technologies and their capabilities have advanced rapidly over the past few years, leading to widespread interest and adoption in the biological and ecological sciences. However, the data infrastructure and computational resources that provide the foundation for AI workflows in ecology and biodiversity science remain fragmented. Moreover, the flow of value remains largely unidirectional: data move from collectors, through repositories and aggregators into models, and finally to research products or management actions. The derived value, such as corrected taxonomies, annotations, quality flags, and insights, rarely returns to source communities. We propose an extension of the original FAIR principles to address the needs of AI-ready infrastructure: the AI-FAIR metadata criteria, which place AI-enabled science in the context of a bidirectional Question–Data–Model triangle. This framework is designed to facilitate large-scale content-based analysis across diverse collections and ensure that the scientists generating the data benefit from the AI-enabled advances built on their data. Alongside these criteria, we introduce an agentic evaluation workflow~\footnote{Code available at: \href{https://github.com/FAIR4AI-Bio/evaluator}{github.com/FAIR4AI-Bio/evaluator}.} that both rates existing datasets on the AI-FAIR and traditional FAIR criteria and provides actionable feedback where critical metadata is missing or incomplete. This modular dataset evaluation tool, paired with a domain-appropriate checklist, translates these principles into a reusable assessment of machine-learning readiness, AI-readiness for specified tasks, traceability, and ethical and CARE-aligned stewardship. Through this approach we aim to create a community-informed system to close the model data loop, so that AI also enriches data to further enable scientific discovery.
\end{abstract}

\section*{From FAIR Data to Bidirectional, AI-Ready Infrastructure}

Artificial Intelligence (AI) is rapidly changing how research in the domains of ecology and biodiversity science is conducted (\hyperref[Pollock]{Pollock et al., 2025}). However, current workflows still rely on data infrastructure built primarily to encourage researcher data contributions rather than to prioritize the machine-readability and cross-dataset interoperability that facilitate data reuse (\hyperref[Crystal]{Crystal-Ornelas et al., 2022}). In both domains, data span species occurrences, images, audio, genomic sequences, sensor observations, field measurements, synthesized species biological information, and environmental covariates (\hyperref[Feng]{Feng et al., 2025}). Heterogeneous communities produce these assets using different terminologies, collecting data at varied spatial and temporal scales, releasing data under diverse licenses and governance frameworks, and storing data in siloed repositories. Fragmented storage, inconsistent formats, limited interoperability, and incomplete scientific context have long presented hurdles for ecological synthesis efforts (\hyperref[OBrien]{O'Brien et al., 2021}), and these barriers have become no more tractable as ecologists develop AI-enabled workflows (\hyperref[Mishra]{Mishra, 2026}). We refer to these persistent obstacles to scientific discovery as the \textbf{FAIR-gap}.

While the original FAIR (Findable, Accessible, Interoperable, Reusable) principles (\hyperref[Wilkinson]{Wilkinson et al., 2016}) remain necessary, they are not sufficient criteria to guide AI-ready infrastructure in the sciences and overcome the FAIR-gap (\hyperref[Huerta]{Huerta et al., 2023}; \hyperref[Clark]{Clark et al., 2026}; \hyperref[Shehu]{Shehu \& Nussinov, 2026}). AI-enabled, data-driven science requires machine-actionable access and interpretation at the appropriate scale. Even a findable, accessible, interoperable, and reusable dataset may require extensive human intervention to determine whether the data are appropriate and permitted for a scientific task, or to connect augmented information (e.g., errors flagged or enhancements) back to the original source.

Extensive global museum digitization efforts, growing community science platform engagement, and increasing accessibility and sensitivity of sensor technologies have produced vast quantities of data faster than can be processed and analyzed through traditional methods. As AI-driven search is increasingly replacing traditional algorithms, FAIR-gapped data risk exclusion from search results, preventing researchers from even considering them. Collecting and curating biodiversity data is labor-intensive; they should not be excluded from consideration because researchers (or AI agents) cannot find or access them, nor should they be misattributed or used for analyses they do not support. The time required to manually sift through these data is rapidly exceeding the capacity of individual researchers. Moreover, the effort, resources, and time that goes into data collection are translated into scientific value and meaning only when those data are used, and attribution is a recognition of those efforts.

Enabling automated assessment of data for scientific use (e.g., using an agentic AI system to perform a broad scale analysis across multiple databases and flag those of interest for manual review) provides a force-multiplier, but this capability requires increased and more consistent documentation of the datasets. Drawing from the FAIR4AI framework (\hyperref[Huerta]{Huerta et al., 2023}) and the FAIR data principles (\hyperref[Wilkinson]{Wilkinson et al., 2016}), we define \textbf{AI-FAIR metadata criteria} as the key to AI-enabled science with bidirectional value. Rather than developing a new prescriptive data specification, we complement ongoing work by Bridge2AI (\hyperref[Clark]{Clark et al., 2026}) and the FAIR\textsuperscript{2} Alliance (\hyperref[Fair2Alliance]{n.d.}). Our framework focuses on foundational metadata components that data providers can integrate into their chosen standards to accelerate scientific discovery and ensure lasting value.

\section*{FAIR4AI-Bio network}\label{fair4ai-bio-network}

The FAIR4AI-Bio network was conceived to address the FAIR-gap by placing the data providers and users at the center of the solutions. This process includes hosting a series of workshops to understand requirements for AI-ready data infrastructure in ecology, biodiversity science, and related fields. The guiding mission through these workshops is to create a community-informed system for unifying biodiversity data at scale, one that promotes trust and encourages the exchange of data. We started by creating an inventory of the platforms and tools used by data producers, providers, managers, and scientists who work with ecological and biodiversity data. We then gathered feedback from the workshop community to identify the challenges and critical blockers encountered with these platforms and tools, and synthesized these into a collection of requirements to address the FAIR-gap: our AI-FAIR framework. This community relies heavily on multimodal data integration for analyses, and addressing this complexity of use-cases will support the design of a robust prototype of the AI-FAIR framework.

\begin{figure}[h]
    \includegraphics[width=\textwidth]{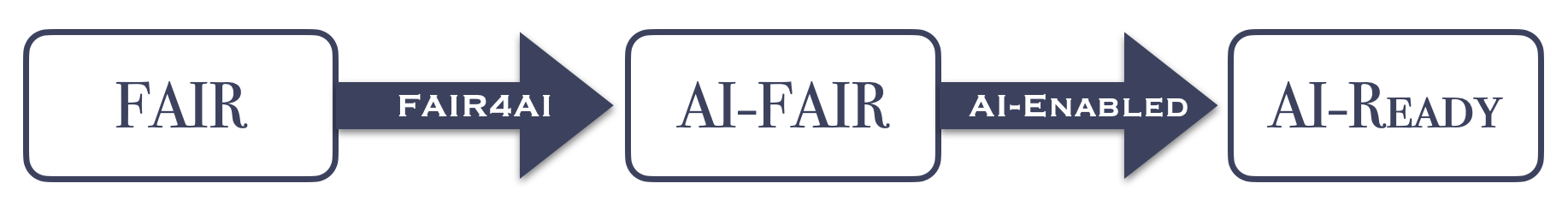}
    \caption{\emph{\textbf{The AI-FAIR continuum.} FAIR principles apply broadly to data without directed downstream use cases. AI-FAIR principles extend traditional FAIR with AI-enabled assessment and processing as a goal. AI-ready is use-case specific, but may be enabled, e.g., through Agentic workflows, when AI-FAIR standards are met.}}
    \label{fig:AI-FAIR-continuum}
\end{figure}

Dataset properties that extend from traditional FAIR principles to AI-FAIR specifically enable machines to use data with minimal human intervention. These include essential metadata to provide the context and controls required for scientifically valid and responsible use at scale. It is impractical to expect data publishers to anticipate the requirements for all possible AI-enabled uses of their data, but AI-FAIR criteria offer guidance for data publishers to provide enhanced metadata that enable more efficient, automated, and accurate inclusion of their data in AI workflows at scale (Figure~\ref{fig:AI-FAIR-continuum}).

\section*{The Question-Data-Model Framework: Roundtripping for Added Value}\label{sec:Q-D-M}

\begin{wrapfigure}{l}{.5\textwidth}
    \includegraphics[width=.47\textwidth]{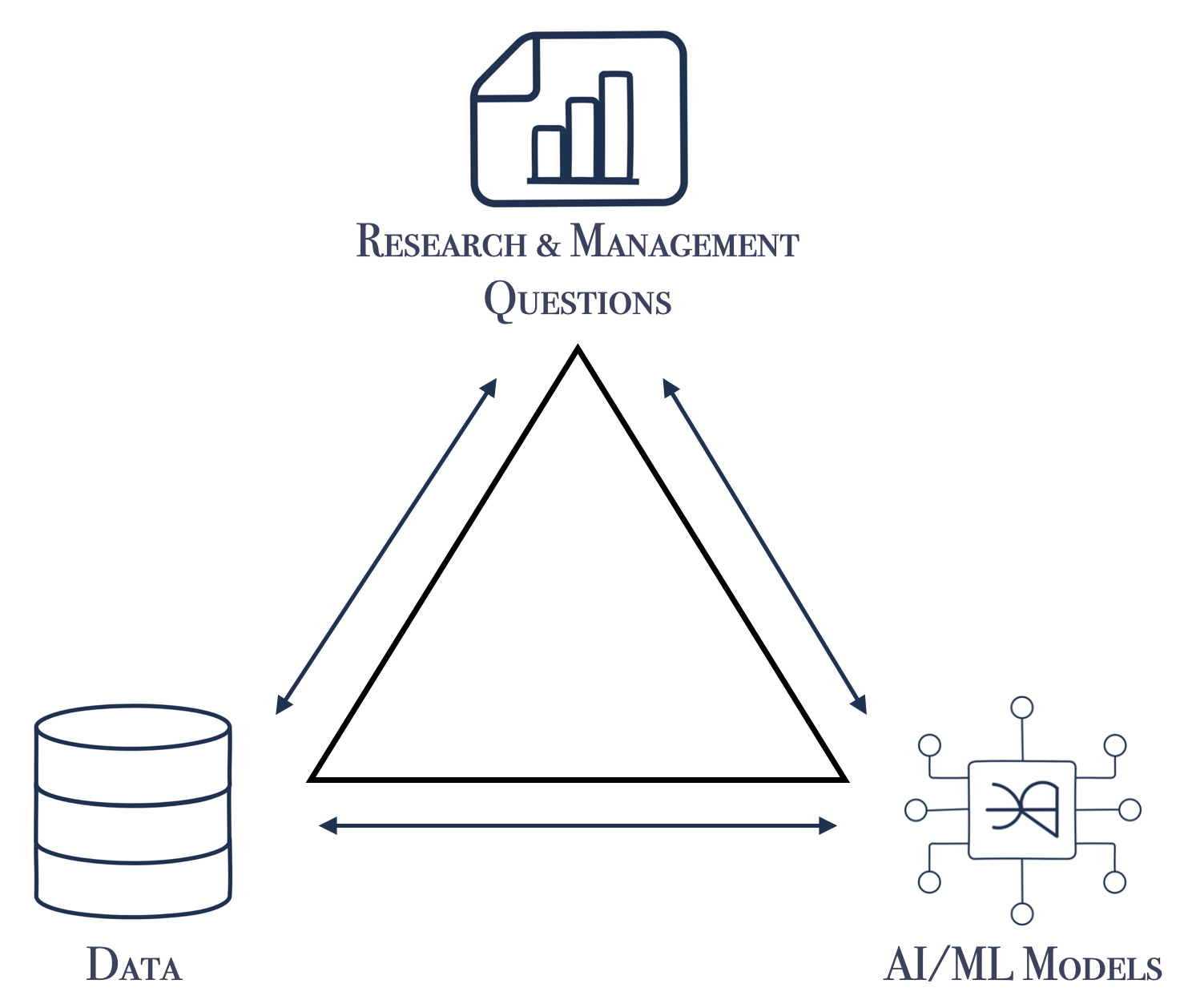}
    \caption{\emph{\textbf{The Question-Data-Model framework.} Questions specify required data and models; models declare capabilities and data needs; data inform and are enriched by questions and models.}}
    \label{fig:Q-D-M-triangle}
\end{wrapfigure}

The AI-FAIR perspective frames AI-enabled science as a Question-Data-Model triangle (Figure~\ref{fig:Q-D-M-triangle}). In a complete system, every edge is bidirectional. Questions specify required data properties and model capabilities.
Models declare their assumptions, capabilities, and data requirements. Data inform both questions and models, while also being enriched by the outputs of scientific interpretation and computation.

However, the data vertex remains mostly unidirectional: data flow from collectors to aggregators, to models, and finally to research products or management actions. At each transition, the provenance chain is often weakened or severed, exacerbating challenges in returning information to data sources. This degradation of provenance increases the risk of \emph{data leakage}, where evaluation data are included in model training. Furthermore, solving this requires additional complex processing---such as perceptual hashing (\hyperref[Zhou]{Zhou et al., 2025})---to disentangle aggregated data that are no longer traceable to their sources. Without context resolution, scientists are limited in the conclusions they can draw from model results, further increasing the likelihood that value created downstream will not return to the people and systems that produced the data used to train and evaluate it.

We posit that data value roundtripping (\hyperref[Larsson]{Larsson et al., 2019}; \hyperref[Groom]{Groom et al., 2022}) must be prioritized in efforts to design distributed AI-ready infrastructure. As AI systems process data---aided by expert validation---enhancing metadata, harmonizing taxonomies, detecting and flagging anomalies, and identifying gaps, they generate intermediate products for which there must be a mechanism to provide these artifacts back to the source repositories. For example, large-scale synthesis in ecology and biodiversity science often involves correcting or aligning taxonomies; identifying bounding boxes for images; detecting and flagging data anomalies; filling gaps in time-series data; assimilating data from other sources, and augmenting labels. Without a provenance chain and standardized metadata detailing the returns desirable by a data collector, all such data augmentation will remain siloed in data archives specific to synthesis projects. Other synthesis efforts will be required to repeat similar processing, with little to no benefit flowing to source repositories, communities, or data producers.

Data sources should be able to incorporate new knowledge with versioning, provenance, and attribution, where appropriate. This, in turn benefits the downstream users and reduces redundant processing. Such an interconnected system has the potential to produce better inputs (e.g., time-series gaps filled, taxonomy aligned, errors corrected) to later workflows, \textbf{creating a living knowledge system that improves through use}.

\section*{A Metadata-First AI-FAIR Framework}\label{ai-fair-framework}

The AI-FAIR framework suggests infrastructure priorities, which can be assessed using the AI-FAIR Checklist. This checklist was initially seeded from the ESIP AI-Readiness Checklist (\hyperref[ESIP]{ESIP Data Readiness Cluster, 2023}) and the Imageomics Dataset Card Checklist (\hyperref[Campolongo]{Campolongo et al., 2026}). Building on these foundations, we engaged workshop participants in the evaluation of the list based on their experiences collecting, hosting, and curating data. Checklist items are those that workshop participants identified as necessary to make data AI-FAIR for scientific research. Each checklist item falls under one of four general metadata categories, which together form the AI-FAIR metadata criteria. We mapped these items to relevant standards including Ecological Metadata Language (\hyperref[Jones]{Jones et al., 2019}), DataCite (\hyperref[DataCite]{DataCite Metadata Working Group, 2026}), Schema.org (\hyperref[SchemaOrg]{Schema.org Community Group, 2026}), and Croissant (\hyperref[Akhtar]{Akhtar et al., 2024}), where the terms exist. This mapping allows the checklist to bridge domain metadata and general AI data documentation. The full, community-approved, schema and mapping will be published with the checklist; it is currently under an open review by the broader biological community to ensure clarity and support. An analogous process could be completed to develop domain-specific keys in other disciplines.

Below, we define the four AI-FAIR metadata criteria, provide examples of checklist categories or items included in them, and what each criterion (or combination of criteria) enables.

\begin{enumerate}
\item
  \textbf{Structural metadata} describe formats, schemas, organization, variables, file relationships, and other values required for machine-readable access, conforming to FAIR standards. These properties determine whether a dataset can be acquired and inserted into a machine-learning pipeline and how data exchange can more readily occur. They support an assessment of \textbf{machine learning (ML) readiness}, but do not by themselves establish scientific validity.
\item
  \textbf{Scientific metadata} describe how observations or datasets were generated and what they mean. For example, a biodiversity use may include sampling design, effort, detectability, sensor placement and calibration, itemized and categorical species data, spatial and temporal resolution, taxonomic treatment and labels, uncertainty, and experimental context. These metadata are often necessary for determining whether data are fit for the intended scientific task. Structural and scientific criteria together support an assessment of \textbf{AI-readiness for particular tasks}.
\item
  \textbf{Provenance metadata} preserve the source chain, identifiers, citations, attribution requirements, checksums, and the environments and software used for processing. Provenance metadata ensures reproducibility of transformations, and paired with scientific requirements, they enable extension beyond an initial intended use and return value to the source data provider. In combination with structural requirements, these criteria support \textbf{traceability}.
\item
  \textbf{Governance metadata} express permissions, restrictions, responsibilities, sensitive-data conditions, and CARE (Collective Benefit, Authority to Control, Responsibility, and Ethics) considerations (\hyperref[Carroll]{Carroll et al., 2020}), in  particular as defined in \emph{IEEE recommended practice for provenance of Indigenous Peoples\textquotesingle{} Data} (\hyperref[IEEE]{2025}). They include information about permission to collect observations and communities, governments, or territories implicated by data use. These criteria support an assessment of \textbf{ethical and CARE-aligned stewardship}.
\end{enumerate}

The checklist further subdivides these criteria into broad domain-agnostic categories of general information, data structure, source data, data processing, data quality, guidance and recommendations, data access, provenance and appropriate usage, and data governance. Within these categories, terms become more domain-specific. Here we provide biodiversity data examples for several elements that workshop participants found particularly important for AI-enabled science. Sampling semantics (such as experimental design, sampling effort, detection processes, instrumentation, calibration, and spatial and temporal scale) help an agent to determine whether a model can generalize beyond observed data. Processing and annotation provenance document steps, such as transformations and methods used for time-series gap filling, outlier handling, anonymization, and labeling, to help prevent data leakage and support reproducible benchmarking. Elements of the checklist pertaining to fitness and limitations ensure information provided by data sources on applicability (e.g., intended tasks, out-of-scope uses) and constraints (e.g., biases in space or time, class imbalances, distinctions between true absences and missing data) are machine-discoverable. Ontological coherence maintains label definitions with clarity and consistency. For instance, in biodiversity use cases the taxonomic backbone, resolution date, synonym handling, and identification uncertainty affect whether records from distributed sources can be combined safely and determine the appropriate process to do so. Finally, components of the checklist are meant to work synergistically. For example, elements pertaining to rights and governance ensure that machine-readable licenses, access tiers, sensitive-species flags, privacy constraints, and CARE-related permissions travel with the data and derivatives to confirm that data are fit for intended tasks.

\section*{From Checklist to Automated Dataset Evaluation}\label{sec:evaluator}

\begin{figure}
    \includegraphics[width=\textwidth]{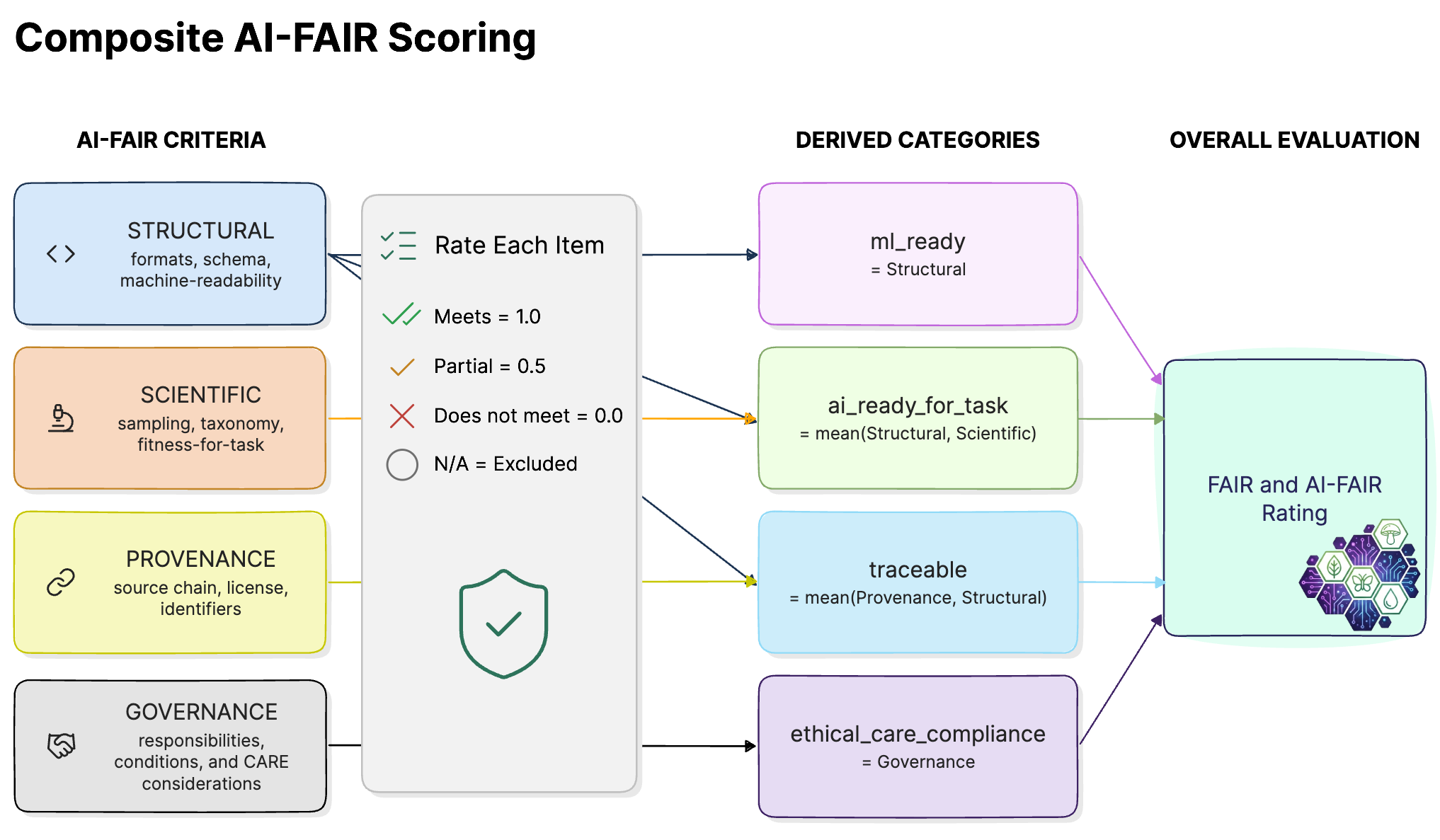}
    \caption{\emph{Checklist ratings are assigned to each item and scored by AI-FAIR metadata criteria: structural, scientific, provenance, and governance. These are then mapped into the four derived categories for algorithmic scoring. Traditional FAIR principle alignment is reported in parallel.}}
    \label{fig:evaluator-workflow}
\end{figure}

We pair the checklist with an automated evaluation workflow (\hyperref[Sokol]{Sokol \& Campolongo, 2026}). The evaluator uses an agent to retrieve available metadata, rate checklist items against a rubric, record evidence and recommendations, and provide FAIR and AI-FAIR scores (Figure~\ref{fig:evaluator-workflow}). The evaluator rates each checklist item as \textbf{meets}, \textbf{partial}, \textbf{does not meet}, or \textbf{not applicable}. Each rating is assigned a set numerical value for scoring, and not applicable is excluded. The agent uses a Python script to average the item scores within each of the four AI-FAIR criteria of metadata, then maps these to their derived categories to compute scores for four components: \textbf{ML-ready}, \textbf{AI-ready for task}, \textbf{Traceable}, and \textbf{Ethical CARE compliance}. These scores then feed into the overall AI-FAIR value for a dataset alongside traditional FAIR dimensions. It is designed such that checklists for other disciplines could be substituted for the ecologically-based one with which we have started.

The separation between evidence gathering and deterministic scoring is important to avoid circularity. An agent may locate metadata and propose ratings, but the aggregation logic is explicit and reproducible rather than estimated by a model. Evaluation output identifies strengths, missing metadata, inaccessible records, and provides concrete recommendations. Critically, these evaluation outputs also retain the evidence supporting each judgment, making the assessment reviewable by data stewards and domain experts.

\section*{Vision}\label{sec:vision}

AI-enabled scientific discovery relies on AI agents being able to do more than simply locate and consume distributed data. Agentic workflows must be able to evaluate dataset suitability for investigating researcher-specified questions, verify usage permissions and responsibilities, and facilitate round-tripping of validated improvements to the data providers. The AI-FAIR paradigm provides a framework for these capabilities through the Question-Data-Model triangle, four complementary metadata criteria (Structural, Scientific, Provenance, Governance), a domain-interchangeable checklist, and a reusable evaluation tool. These components are generalizable beyond biodiversity data through substitution of discipline-specific values or modules into the checklist. The key is to map discipline-specific categories to their associated AI-FAIR criteria.

Data publishers participating in the FAIR4AI-Bio project, including the National Ecological Observatory Network (NEON) and NatureServe, have begun piloting the evaluation tool to establish baseline AI-FAIR scores for their data. These quantifiable metrics measure progress toward the goal of providing AI-FAIR data, as defined by the multidisciplinary users of these data. Here we have the opportunity to engage the communities that use and produce the data in assessing and setting standards that better enable their workflows and scientific or management pursuits. Especially as AI is becoming increasingly integrated in those efforts, the AI-FAIR framework preserves disciplinary context while enabling cross-domain infrastructure. It aims to support agentic workflows without disconnecting computation from human expertise and responsibility. Most importantly, it provides a roadmap to replace one-way extraction with a roundtrip of data, evidence, attribution, and value, so that the scientists generating the data fully participate in the benefits of these AI-enabled advances. Such a fabric would not be a passive network of repositories but a distributed, governed, and continuously improving scientific knowledge system, collectively shaped by the community.

\section*{Acknowledgments}\label{acknowledgments}

The authors would like to thank all the participants of \emph{FAIR4AI-Bio Workshop 1: AI-Ready Ecology and Biodiversity Data Infrastructure for Open Science}, for their insightful conversations and feedback on our AI-FAIR checklist. This work is supported by the U.S. National Science Foundation awards 2531922, 2531924, 2531925, 2600200 ``Collaborative Research: Disciplinary Improvements: AI-ready Ecology and Biodiversity Data Infrastructure for Open Science'' and 2118240 "HDR Institute: Imageomics: A New Frontier of Biological Information Powered by Knowledge-Guided Machine Learning." This work was also supported by the U.S. National Science Foundation Award DBI-2438843 ``Center: National Synthesis Center for Organismal Resilience to Environmental Change (NSCORE)''. SR was additionally supported by Hatch project Award \#MEO-022425 from the U.S. Department of Agriculture's National Institute of Food and Agriculture. This material is based in part upon work supported by the National Ecological Observatory Network (NEON), a program sponsored by the U.S. National Science Foundation and operated under cooperative agreement by Battelle.

\section*{References}\label{references}

\aparef\phantomsection\label{Akhtar}{}Akhtar, M., Benjelloun, O., Conforti, C., Foschini, L., Gijsbers, P., Giner-Miguelez, J., Goswami, S., Jain, N., Karamousadakis, M., Krishna, S., Kuchnik, M., Lesage, S., Lhoest, Q., Marcenac, P., Maskey, M., Mattson, P., Oala, L., Oderinwale, H., Ruyssen, P.,\ldots. Zhang, L. (2024). Croissant: A Metadata Format for ML-Ready Datasets. In Globerson, A., Mackey, L., Belgrave, D., Fan, A., Paquet, U., Tomczak, J., \& Zhang, C. (Eds.), Advances in Neural Information Processing Systems (Vol. 37, pp. 82133--82148). Retrieved from \url{https://proceedings.neurips.cc/paper\_files/paper/2024/file/9547b09b722f2948ff3ddb5d86002bc0-Paper-Datasets\_and\_Benchmarks\_Track.pdf}

\aparef\phantomsection\label{Campolongo}{}Campolongo, E., Thompson, M., Zhang, N., Duan, Y., Bradley, J., Zarubiieva, I., Taylor, G., \& Lapp, H. (2026). Collaborative Distributed Science Guide (Version 2.0.0) {[}Computer software{]}. Zenodo. \url{https://doi.org/10.5281/zenodo.21301967}

\aparef\phantomsection\label{Carroll}{}Carroll, S. R., Garba, I.,
Figueroa-Rodríguez, O. L., Holbrook, J., Lovett, R., Materechera, S., Parsons, M., Raseroka, K., Rodriguez-Lonebear, D., Rowe, R., Sara, R., Walker, J. D., Anderson, J., \& Hudson, M. (2020). The care principles for indigenous data governance. \emph{Data Science Journal}, \emph{19}, 43. \url{https://doi.org/10.5334/dsj-2020-043}

\aparef\phantomsection\label{Clark}{}Clark, T., Caufield, H., Parker, J. A., Al Manir, S., Amorim, E., Eddy, J., Gim, N., Gow, B., Goar, W., Hansen, J. N., Harris, N., Hermjakob, H., Joachimiak, M., Jordan, G., Lee, I.-H., McWeeney, S. K., Nebeker, C., Nikolov, M., Ratcliffe, S. J., \ldots{} Munoz-Torres, M. C. (2026). \emph{AI-Readiness Criteria for Biomedical Data}. bioRxiv. \url{https://doi.org/10.1101/2024.10.23.619844}

\aparef\phantomsection\label{Crystal}{}Crystal-Ornelas, R., Varadharajan, C., O'Ryan, D., Beilsmith, K., Bond-Lamberty, B., Boye, K., Burrus, M., Cholia, S., Christianson, D. S., Crow, M., Damerow, J., Ely, K. S., Goldman, A. E., Heinz, S. L., Hendrix, V. C., Kakalia, Z., Mathes, K., O'Brien, F., Pennington, S. C., \ldots{} Agarwal, D. A. (2022). Enabling fair data in Earth and environmental science with community-centric (meta)data reporting formats. \emph{Scientific Data}, \emph{9}(1). \url{https://doi.org/10.1038/s41597-022-01606-w}

\aparef\phantomsection\label{DataCite}{}DataCite Metadata Working Group. (2026). \emph{DataCite Metadata Schema for the Publication and Citation of Research Data and Other Research Outputs}. (Version 4.7). DataCite e.V. \url{https://doi.org/10.14454/28a4-kd32}

\aparef\phantomsection\label{ESIP}{}ESIP Data Readiness Cluster. (2023). \emph{Checklist to Examine AI-readiness for Open Environmental Datasets} (Version 1.0). Earth Science Information Partners. Retrieved June 30, 2025, from \url{https://github.com/ESIPFed/data-readiness}

\aparef\phantomsection\label{Fair2Alliance}{}\emph{Fair\textsuperscript{2}} \emph{Alliance}. (n.d.). FAIR\textsuperscript{2}. Retrieved October 14, 2026, from \url{https://fair2.ai/}

\aparef\phantomsection\label{Feng}{}Feng, X., Smith, A. B., Boyle, B., Chen, X., Enquist, B. J., Gallagher, R., Hammock, J., Ho, J. C., Lien, A. M., Maitner, B., Sokol, E. R., Soltis, P. S., Wenk, E. H., Willoughby, A., \& Park, D. S. (2025). The next stage of Biodiversity Informatics: Community-driven synthesis and integration of biodiversity databases. \emph{BioScience}, \emph{75}(11), 913--925. \url{https://doi.org/10.1093/biosci/biaf112}

\aparef\phantomsection\label{Groom}{}Groom, Q., Bräuchler, C., Cubey, R. W. N., Dillen, M., Huybrechts, P., Kearney, N., Klazenga, N., Leachman, S., Paul, D. L., Rogers, H., Santos, J., Shorthouse, D. P., Vaughan, A., von Mering, S., \& Haston, E. M. (2022). The disambiguation of people names in Biological Collections. \emph{Biodiversity Data Journal}, \emph{10}. \url{https://doi.org/10.3897/bdj.10.e86089}

\aparef\phantomsection\label{Huerta}{}Huerta, E. A., Blaiszik, B., Brinson, L. C., Bouchard, K. E., Diaz, D., Doglioni, C., Duarte, J. M., Emani, M., Foster, I., Fox, G., Harris, P., Heinrich, L., Jha, S., Katz, D. S., Kindratenko, V., Kirkpatrick, C. R., Lassila-Perini, K., Madduri, R. K., Neubauer, M. S., \ldots{} Zhu, R. (2023). Fair for AI: An Interdisciplinary and International Community Building Perspective. \emph{Scientific Data}, \emph{10}(1). \url{https://doi.org/10.1038/s41597-023-02298-6}

\aparef\phantomsection\label{IEEE}{}IEEE recommended practice for provenance of Indigenous Peoples' Data. (2025). \emph{IEEE Standards 2890‐2025}, 1--25. \url{https://doi.org/10.1109/ieeestd.2025.11302960}

\aparef\phantomsection\label{Jones}{}Jones, M.B., O'Brien, M., Mecum, B., Boettiger, C., Schildhauer, M., Maier, M., Whiteaker, T., Earl, S., Chong, S. (2019). \emph{Ecological Metadata Language} (Version 2.2.0). KNB Data Repository. \url{https://doi.org/10.5063/F11834T2}

\aparef\phantomsection\label{Larsson}{}Larsson, A., Ånäs, S., and Zeinstra, M. (2019). \emph{Data Roundtripping: A new frontier for GLAM-Wiki collaborations}. Wikimedia Diff. \url{https://upload.wikimedia.org/wikipedia/commons/e/e8/Wikimedia_Commons_Data_Roundtripping_-_Final_report.pdf}

\aparef\phantomsection\label{Mishra}{}Mishra, G. (2026). Ecology is not yet ready for AI---and why that matters. \emph{Proceedings of the National Academy of Sciences}, \emph{123}(22). \url{https://doi.org/10.1073/pnas.2533976123}

\aparef\phantomsection\label{OBrien}{}O'Brien, M., Smith, C. A., Sokol, E. R., Gries, C., Lany, N., Record, S., \& Castorani, M. C. N. (2021). EcocomDP: A flexible data design pattern for Ecological Community Survey Data. \emph{Ecological Informatics}, \emph{64}, 101374. \url{https://doi.org/10.1016/j.ecoinf.2021.101374}

\aparef\phantomsection\label{Pollock}{}Pollock, L. J., Kitzes, J., Beery, S., Gaynor, K. M., Jarzyna, M. A., Mac Aodha, O., Meyer, B., Rolnick, D., Taylor, G. W., Tuia, D., \& Berger-Wolf, T. (2025). Harnessing artificial intelligence to fill global shortfalls in Biodiversity Knowledge. \emph{Nature Reviews Biodiversity}, \emph{1}(3), 166--182. \url{https://doi.org/10.1038/s44358-025-00022-3}

\aparef\phantomsection\label{SchemaOrg}{}Schema.org Community Group (2026, March 19). \emph{Schema.org is a set of extensible schemas that enables webmasters to embed structured data on their web pages for use by search engines and other applications.}. (Version 30.0). \url{https://schema.org/}

\aparef\phantomsection\label{Shehu}{}Shehu, A., \& Nussinov, R. (2026). Four questions for AI-ready biological data. \emph{Nature Machine Intelligence}, \emph{8}(7), 1017--1019. \url{https://doi.org/10.1038/s42256-026-01270-4}

\aparef\phantomsection\label{Sokol}{}Sokol, E. R., \& Campolongo, E. (2026). AI-FAIR Data Evaluator (Version 0.2.1) {[}Computer software{]}. \url{https://doi.org/10.5281/zenodo.23744163}

\aparef\phantomsection\label{Wilkinson}{}Wilkinson, M. D., Dumontier, M., Aalbersberg, Ij. J., Appleton, G., Axton, M., Baak, A., Blomberg, N., Boiten, J.-W., da Silva Santos, L. B., Bourne, P. E., Bouwman, J., Brookes, A. J., Clark, T., Crosas, M., Dillo, I., Dumon, O., Edmunds, S., Evelo, C. T., Finkers, R., \ldots{} Mons, B. (2016). The Fair Guiding Principles for Scientific Data Management and Stewardship. \emph{Scientific Data}, \emph{3}(1). \url{https://doi.org/10.1038/sdata.2016.18}

\aparef\phantomsection\label{Zhou}{}Zhou, Y., Li, X., Xiong, C., Yao, H., \& Qin, C. (2025). A survey of perceptual hashing for Multimedia. ACM Transactions on Multimedia Computing, Communications, and Applications, 21(7), 1--28. \url{https://doi.org/10.1145/3727880}

\end{document}